\documentclass[conference]{IEEEtran}
\IEEEoverridecommandlockouts
\usepackage{cite}
\usepackage{amsmath,amssymb,amsfonts}
\usepackage{algorithmic}
\usepackage{graphicx}
\usepackage{textcomp}
\usepackage{xcolor}
\usepackage{booktabs}
\usepackage{multirow}

\usepackage{float}

\usepackage{tikz}
\usetikzlibrary{shapes.geometric, arrows, positioning}

\tikzstyle{startstop} = [rectangle, rounded corners, minimum width=3cm, minimum height=1cm, text centered, draw=black, fill=red!30]
\tikzstyle{process} = [rectangle, minimum width=3cm, minimum height=1cm, text centered, draw=black, fill=orange!30]
\tikzstyle{arrow} = [thick,->,>=stealth]
\def\BibTeX{{\rm B\kern-.05em{\sc i\kern-.025em b}\kern-.08em
    T\kern-.1667em\lower.7ex\hbox{E}\kern-.125emX}}

\usepackage{fancyhdr}  % Add this to the preamble
\fancypagestyle{firstpage}{
  \fancyhf{} % Clear all header and footer fields
  \fancyhead[L]{\small 2025 International Conference on Quantum Photonics, Artificial Intelligence, and Networking (QPAIN) \\ 31 July – 2 August 2025, Rangpur, Bangladesh}
  \fancyfoot[L]{\small 979-8-3315-9694-1/25/\$31.00 \copyright2025 IEEE}

}

\begin{document}

\title{Strengthening False Information Propagation Detection: Leveraging SVM and Sophisticated Text Vectorization Techniques in Comparison to BERT\\

{\footnotesize \textsuperscript{}}
\thanks{}
}

\author{
\IEEEauthorblockN{Ahmed Akib Jawad Karim}
\IEEEauthorblockA{\textit{Dept. of CSE} \\
\textit{BRAC University}\\
Dhaka, Bangladesh \\
akibjawaad@gmail.com}
\and
\IEEEauthorblockN{Kazi Hafiz Md Asad}
\IEEEauthorblockA{\textit{Dept. of ECE} \\
\textit{North South University}\\
Dhaka, Bangladesh \\
kaziasad061@gmail.com}
\and
\IEEEauthorblockN{Aznur Azam}
\IEEEauthorblockA{\textit{Dept. of CSE} \\
\textit{BAUST}\\
Saidpur, Bangladesh \\
aznurazam2@gmail.com
}
}

\maketitle
\thispagestyle{firstpage}

\begin{abstract}
The rapid spread of misinformation, particularly through online platforms, underscores the urgent need for reliable detection systems. This study explores the utilization of machine learning and natural language processing, specifically Support Vector Machines (SVM) and BERT, to detect fake news. We employ three distinct text vectorization methods for SVM: Term Frequency Inverse Document Frequency (TF-IDF), Word2Vec, and Bag of Words (BoW), evaluating their effectiveness in distinguishing between genuine and fake news. Additionally, we compare these methods against the transformer large language model, BERT. Our comprehensive approach includes detailed preprocessing steps, rigorous model implementation, and thorough evaluation to determine the most effective techniques. The results demonstrate that while BERT achieves superior accuracy with 99.98\% and an F1-score of 0.9998, the SVM model with a linear kernel and BoW vectorization also performs exceptionally well, achieving 99.81\% accuracy and an F1-score of 0.9980. These findings highlight that, despite BERT's superior performance, SVM models with BoW and TF-IDF vectorization methods come remarkably close, offering highly competitive performance with the advantage of lower computational requirements.

\end{abstract}

% \begin{abstract}
% The rapid spread of misinformation, particularly through online platforms, underscores the urgent need for reliable detection systems. This study explores the utilization of machine learning and natural language processing, specifically Support Vector Machines (SVM) and BERT, to detect news that are fake. We employ three distinct text vectorization methods for SVM: Term Frequency Inverse Document Frequency (TF-IDF), Word2Vec, and Bag of Words (BoW), evaluating their effectiveness in distinguishing between genuine and fake news. Additionally, we compare these methods against the transformer large language model BERT. Our comprehensive approach includes detailed preprocessing steps, rigorous model implementation, and thorough evaluation to determine the most effective techniques. The results demonstrate that while BERT achieves superior accuracy, the SVM models with BoW and TF-IDF vectorization methods come remarkably close, offering highly competitive performance with the advantage of lower computational requirements.

% \end{abstract}


\section{Introduction}
The expansion of fake news has emerged as a significant threat to public opinion and societal stability. Rapid dissemination of misinformation via social media and online platforms has led to crises affecting political decisions, public health, and social harmony. Research indicates that fake news spreads significantly faster and more widely than true news, often leading to real-world consequences. For example, during the 2016 U.S. Presidential Election, fake news stories were more engaging on Facebook than real news. According to the Pew Research Center, 64\% of Americans feel false news has seriously misled them about current affairs. The massive amount of data produced daily renders conventional hand fact-checking methods insufficient. Apart from confusing the people, false news can motivate violence and compromise the democratic framework. Fake news spread over WhatsApp has been linked to political unrest and societal instability in Bangladesh as well as mob violence and lynchings in India. During the COVID-19 pandemic, false information about the virus and vaccinations seriously threatened public health. These events show how crucial it is to have trustworthy mechanisms in place to spot false news.
Consequently, automated systems that can identify bogus news utilizing artificial intelligence (AI) and machine learning (ML) techniques are in more demand. By fast analyzing large datasets and spotting trends implying the inclusion of false information, artificial intelligence (AI) tools can dramatically increase the accuracy and effectiveness of spotting fake news.

This work provides a fresh natural language processing method based on Support Vector Machines (SVM) combined with three text vectorizing techniques and a transformer LLM model—more especially, BERT-base—that precisely labels news items as either fake or real. By means of Term Frequency-Inverse Document Frequency (TF-IDF), Word2Vec, and Bag of Words (BoW) approaches, the paper aims to raise the accuracy and dependability of false news detection systems. By contrasting these vectorizing techniques with BERT and proving the effectiveness of SVM classifiers, this study aids to the preservation of information integrity and the fight against false information in the digital world.

The research presents significant contributions and originality in the following aspects:

\begin{itemize}

\item It compares BoW, TF-IDF, and Word2Vec vectorization methods for SVM classifiers, providing insights into their performance for fake news detection. 
\item It analyzes the marginal benefits of the RBF kernel over the linear kernel in SVM for fake news detection. The research utilizes a preprocessed version of the ISOT fake news dataset, ensuring the practical relevance and applicability of the findings.
\item The paper contrasts SVM with the BERT base model, highlighting their respective advantages and limitations, particularly in resource-constrained environments. 

\end{itemize}

\section{Related Work}

Several studies have explored various approaches for fake news detection. Shu et al. \cite{shu2017fake} reviewed data mining methods and challenges involved in detecting fake news, emphasizing the need for comprehensive techniques. Zhang et al. \cite{zhang2018fake} investigated linguistic features for fake news detection and proposed a composite model combining LSTM and CNN, achieving an accuracy of 92.4\%. Rashkin et al. \cite{rashkin2017truth} utilized linguistic cues, showing that stylistic features can indicate deceptive content. Pérez-Rosas et al. \cite{perez2018automatic} explored lexical, syntactic, and semantic features, reporting an F1 score of 0.78, highlighting the importance of feature diversity. Wang \cite{wang2017liar} introduced the LIAR dataset and developed several machine learning models, with logistic regression achieving an accuracy of 24.7\%, underscoring the challenges in fake news identification. Ruchansky et al. \cite{ruchansky2017csi} presented the CSI model, integrating user behavior analysis with content analysis, achieving an accuracy of 89.2\%. Dhruv Khattar et al. \cite{khattar2019mvae} developed a multi-modal variational autoencoder integrating text and image features, resulting in an accuracy of 94.1\%. Bhardwaj et al. \cite{bhardwaj2020fake} proposed a composite approach combining CNN and BiLSTM, achieving an accuracy of 96.7\%. Shu and Bin Wu \cite{shu2021temporal} introduced a time-evolving graph neural network, capturing the dynamic nature of news dissemination, with notable performance improvements. Alnabhan and Branco \cite{alnabhan2024evaluating} tested deep learning models for cross-domain false news identification. Karim et al.\cite{karim2024largermodelsyieldbetter} explore the effectiveness of larger BERT-based models in streamlining the classification of ADHD-related concerns through knowledge distillation, demonstrating the potential of simplified models without sacrificing performance. Their work highlights the adaptability of BERT for nuanced tasks, which could inform advancements in domains like fake news detection by leveraging sophisticated model architectures and text representations. Combining word embeddings with linguistic features, Verma et al. \cite{verma2021welfake} presented WELFake, which greatly raised accuracy and F1-score. By means of a benchmark for assessing the efficacy of several approaches, Kanavos et al. \cite{kanavos2023comparative} performed a comparative analysis of machine learning algorithms and text vectorizing techniques for false news identification. Using SVM classifiers with various text vectorizing techniques—including TF-IDF, Word2Vec, and BoW—this work expands on these foundations to identify the best efficient text representation method for spotting bogus news.

By using SVM classifiers with various text vectorizing techniques, including TF-IDF, Word2Vec, and BoW, this work builds on these foundations to ascertain the most efficient text representation method for spotting bogus news. This thorough comparison seeks to improve the dependability and accuracy of false news detecting systems.

\section{Methodology}
This part covers in great detail the approaches to create the false news detecting mechanism. Data collecting, preprocessing, text vectorizing, and Support Vector Machine (SVM) and BERT based categorization is part of the method.

\subsection{Dataset and Preprocessing}

Designed by the Information Security and Object Technology (ISOT) Research Group at the University of Victoria \cite{ISOT}, Kaggle's "Fake and Real News Dataset" \cite{FakeNewsDataset} provides the dataset used in this research effort. Labeled as either true or false, this dataset comprises thorough news items of political news, world news, government news, Middle-Eastern news, US news. While the bogus news items came from many websites highlighted by Politifact and Wikipedia, the actual news items were gathered from Reuters.com. Our preprocessed data contains 21,477 real news items and almost 23,421 false news items. Fake news items are denoted with a "0," while actual news items have a "1." The dataset offers a large base for study since it covers a wide spectrum of news subjects. To guarantee the best quality for later study, the data is carefully cleaned and preprocessed. The distribution of real and false news items in both the training and test datasets is "Train Fake" with 18,796 articles, "Train Real" with 17,121 pieces, "Test Fake" with 4,624 posts, and "Test Real" with 4,356 articles.

Preprocessing involves several steps to clean and prepare the text data for analysis. Initially, the text is cleaned by removing stopwords and unwanted characters and then tokenized. The process utilizes the \texttt{gensim} library for tokenization, which breaks down the text into individual words while removing stopwords and words shorter than three characters. Specifically, the steps include removing stopwords using the NLTK library's predefined list of English stopwords and tokenizing the text with the \texttt{gensim.utils.simple\_preprocess} function that converts the text into a list of lowercase tokens and removes punctuations and words shorter than three characters, and finally reassembles the tokenized lists back into cleaned text strings. For BERT training, the dataset was tokenized using BertTokenizer from pre trained BERT-base-uncased model.

\begin{figure}[htbp]
\centering
\includegraphics[width=\linewidth]{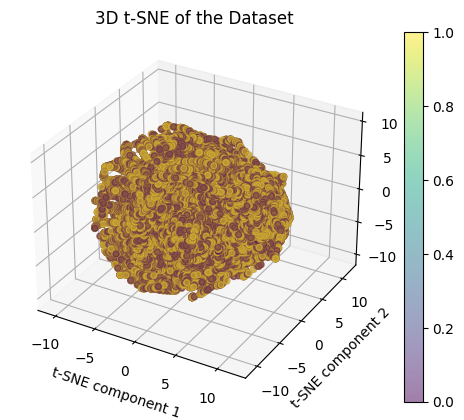}
\caption{3D t-SNE of the Dataset}
\label{fig:3d_tsne}
\end{figure}

The 3D t-SNE plot presented in \textbf{Figure \ref{fig:3d_tsne}} offers a more nuanced and comprehensive view of the dataset by adding an additional dimension to the visualization. This three-dimensional perspective enhances our ability to perceive the complex relationships and interactions between data points that might not be as apparent in the 2D plot. The colors in this 3D plot represent different clusters, each potentially corresponding to distinct groups of fake or real news articles. This clustering is indicative of the inherent structure within the dataset and supports the hypothesis that the features used in our model are effective in capturing the distinctions between the two classes. By examining these clusters, we can gain deeper insights into the data's distribution and the model's potential to classify new, unseen data accurately. The 3D t-SNE plot thus not only validates the separability observed in the 2D plot but also highlights the complex, multi-dimensional nature of the data, reinforcing the robustness of our chosen vectorization and classification techniques.

\subsection{Text Vectorization}
Text vectorization transforms textual data into numerical vectors. This study employs three techniques: TF-IDF, Word2Vec and BoW.

\subsubsection{TF-IDF}
This reflects the importance of terms within a document relative to a corpus. This technique mitigates the limitations of BoW by considering the inverse document frequency, thus reducing the weight of common terms and highlighting unique terms. The mathematical representation is:
\[
\text{TF-IDF}(x,d) = \text{TF}(x,d) \times \log \left(\frac{M}{m_t}\right)
\]
where \( M \) represents the total number of documents, and \( m_t \) denotes the number of documents that include the term \( x \).

\subsubsection{Word2Vec}
Word2Vec generates dense vector representations for words by training on a large corpus. This technique captures semantic relationships, allowing words with similar meanings to have similar vector representations. This study uses the continuous bag-of-words, also known as CBOW architecture which is used to predict words within a context window, thus embedding semantic meaning into the vectors.

\subsubsection{Bag of Words (BoW)}
BoW converts text into a matrix of token counts. This method, while straightforward, is effective for capturing word frequency, though it may miss context and semantics. We apply this method to establish a baseline for comparison with more sophisticated techniques.

\subsection{Classification using SVM}
Support Vector Machines (SVMs) are utilized for news article classification using both linear and radial basis function (RBF) kernels. 

\subsubsection{SVM with Linear Kernel}
The linear kernel is employed to classify the vectorized text. The objective of the SVM is to identify a hyperplane that optimally separates data points into different classes. Considering a set of training examples \((x_i, y_i)\), where \(x_i\) represents the feature vector and \(y_i\) denotes the class label, the SVM solves the optimization problem:

\begin{equation}
\min_{\mathbf{w}, h} \frac{1}{2} \mathbf{w}^T \mathbf{w} + R \sum_{j=1}^{n} \max(0, 1 - y_i (\mathbf{w} \cdot x_j + h))
\end{equation}

where \(\mathbf{w}\) is the weight vector, \(h\) is the bias, \(R\) is the regularization parameter, and \(n\) dictates the number of training examples.

\subsubsection{SVM with RBF Kernel}
Additionally utilized for vectorized text classification is the RBF kernel. Formulated as: the Radial Basis Function (RBF) kernel is:

\begin{equation}
%F(x_a, x_b) = \exp \left( -\alpha \| x_a - x_b \|^2 \right)
F(x_a, x_b) = \exp \left( -\alpha \| (x_a - x_b)^2 \| \right)
\end{equation}

where the influence of a single training example is determined by \(\alpha\).

\begin{figure}[htbp]
\centering
\begin{tikzpicture}[node distance=2cm, auto]

% Nodes
\node (start) [startstop] {Start};
\node (collect) [process, below of=start] {Data Collection};
\node (clean) [process, below of=collect] {Data Cleaning};
\node (preprocess) [process, below of=clean] {Preprocessing};
\node (vectorize) [process, below of=preprocess] {Text Vectorization};
\node (split) [process, right of=collect, xshift=3cm] {Train-Test Split};
\node (classify) [process, below of=split] {Training on SVM Kernals \& BERT};
\node (evaluate) [process, below of=classify] {Model Evaluation};
\node (stop) [startstop, below of=evaluate] {End};

% Arrows
\draw [arrow] (start) -- (collect);
\draw [arrow] (collect) -- (clean);
\draw [arrow] (clean) -- (preprocess);
\draw [arrow] (preprocess) -- (vectorize);
\draw [arrow] (vectorize.east) -| ([xshift=0.5cm]vectorize.east) |- (split);
\draw [arrow] (split) -- (classify);
\draw [arrow] (classify) -- (evaluate);
\draw [arrow] (evaluate) -- (stop);

\end{tikzpicture}
\caption{Workflow Diagram for Fake News Detection}
\label{fig:workflow}
\end{figure}
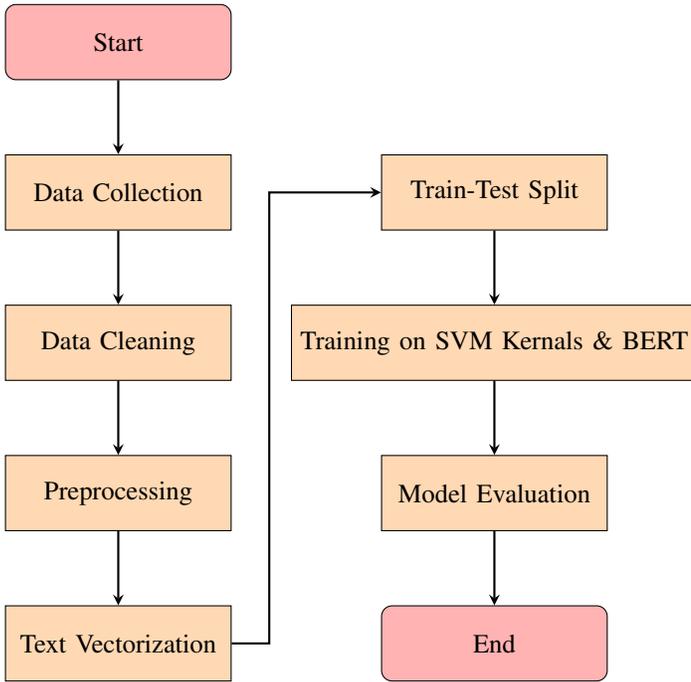

Figure \ref{fig:workflow} shows the method for creating a fictitious news detecting system. Data collecting from several sources comes first, then data cleansing to eliminate extraneous material. Then preprocessing uses stemming, stopword elimination, and tokenizing to change the text. Bag of Words (BoW), TF-IDF, or Word2Vec help to vectorize the text into numerical representations. Training and testing sets separate the data; models including Support Vector Machines (SVM) and BERT are trained to categorize bogus and legitimate news items. To guarantee the efficiency of the detection system, model evaluation is lastly conducted evaluating accuracy, precision, recall, and F1 score.

% \begin{figure}[htbp]
% \centering
% \includegraphics[width=\linewidth]{workflow.png}
% \caption{Confusion Matrix (Linear Kernel)}
% \label{fig:workflow}
% \end{figure}

%--------------------------------------------------------------------
\section{Result Analysis}

The results show that the SVM classifiers—especially with BoW and TF-IDF vectorizing methods—perform remarkably well in spotting bogus news. Demonstrating its efficiency for this work, the BoW method with a linear kernel produced the best accuracy and F1 score. Lastly, the training and analysis of BERT base are also demonstrated.

\subsection{Performance of SVM with TF-IDF}
The TF-IDF vectorizing method performed an accuracy of 99.52\% and an F1 score of 0.9949, the SVM classifier running a linear kernel These findings, shown in Table \ref{tab:combined_results}, show that TF-IDF is likewise a useful technique for text vectorizing in false news identification.

\begin{figure}[htbp]
\centering
\includegraphics[width=\linewidth]{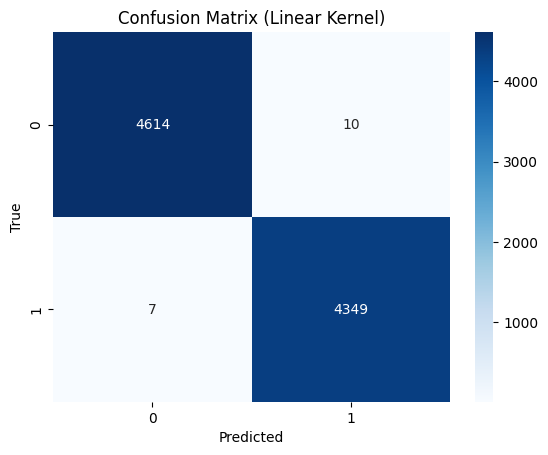}
\caption{Confusion Matrix (Linear Kernel)}
\label{fig:svm_confusionMatrix}
\end{figure}

\textbf{Figure \ref{fig:svm_confusionMatrix}}: The confusion matrix shows how well the classifier properly and erroneously classified events.

\subsection{Performance of SVM with Word2Vec}
The Word2Vec vectorization technique resulted in a lower performance compared to BoW and TF-IDF. The SVM classifier with a linear kernel attained an accuracy of 96.54\% and an F1-score of 0.9644. Although Word2Vec captures semantic meanings of words, it appears that for this particular task, BoW and TF-IDF are more effective. The outcomes are detailed in Table \ref{tab:combined_results}.

\subsection{Performance of SVM with Bag of Words (BoW)}
Employing the Bag of Words (BoW) vectorization technique, the SVM classifier with a linear kernel attained an accuracy of 99.81\% and an F1-score of 0.9980. These high metrics demonstrate that the model excelled in differentiating between fake and real news articles. The detailed results are presented in Table \ref{tab:combined_results}.

\begin{figure}[htbp]
\centering
\includegraphics[width=\linewidth]{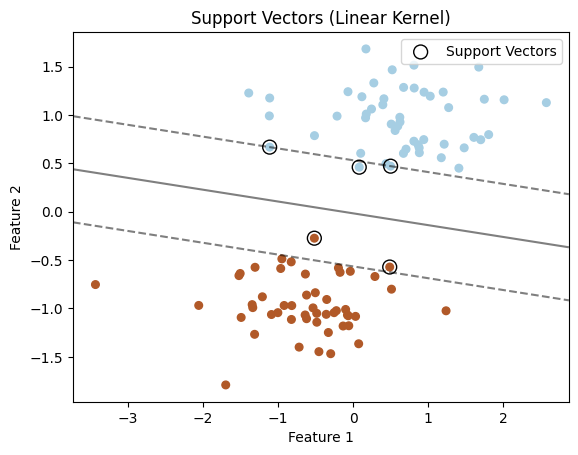}
\caption{Support Vectors (Linear Kernel)}
\label{fig:support_vectors}
\end{figure}

\begin{figure}[htbp]
\centering
\includegraphics[width=\linewidth]{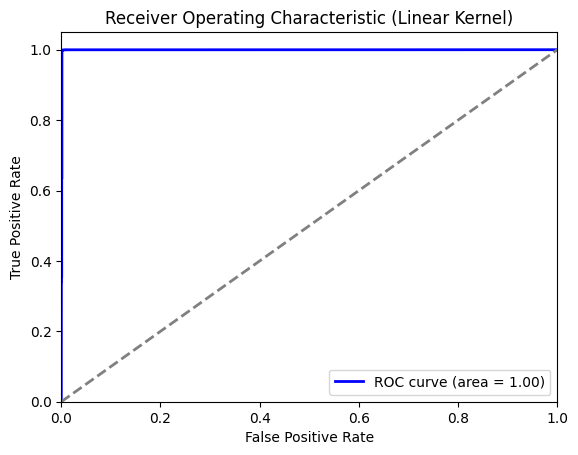}
\caption{Receiver Operating Characteristic (Linear Kernel)}
\label{fig:roc_curve}
\end{figure}

In \textbf{Figure \ref{fig:support_vectors}}, we can observe how the plots illustrate the support vectors identified by the SVM with a linear kernel. These vectors are critical as they define the optimal hyperplane that separates the fake news from the real news in the feature space. Moreover, in \textbf{Figure \ref{fig:roc_curve}}, the ROC curve shows the true positive rate against the false positive rate for different threshold values. The area under the curve (AUC) of 1.0 indicates perfect classification by the SVM model with a linear kernel.

\subsection{Performance of SVM with Radial Basis Function Kernel}
Additionally, the SVM classifier with a radial basis function (RBF) kernel was evaluated for all three vectorization techniques. The results show a slight improvement over the linear kernel in some cases.

\subsubsection{BoW with RBF Kernel}
The SVM classifier utilizing the BoW technique combined with an RBF kernel attained an accuracy of 99.62\% and an F1-score of 0.9961, as illustrated in Table \ref{tab:combined_results}. These results highlight the effectiveness of the RBF kernel in improving the model's performance. The slight improvement in accuracy and F1-score indicates that the RBF kernel can better capture the non-linear relationships in the data, enhancing the classifier's ability to distinguish between fake and real news articles.

\subsubsection{TF-IDF with RBF Kernel}

Table \ref{tab:combined_results} shows the SVM classifier employing TF-IDF vectorization with an RBF kernel obtaining an accuracy of 99.31\% and an F1-score of 0.9928. This shows a great degree of recall and accuracy, therefore proving the strength of the TF-IDF and RBF kernel mix. The findings imply that this method provides a reliable means of differentiating between fake and legitimate news sources since it successfully catches the subtleties in the textual data.

\subsubsection{Word2Vec with RBF Kernel}
As shown in Table \ref{tab:combined_results}, ultimately the SVM classifier using Word2Vec with an RBF kernel achieved an accuracy of 97.75\% and an F1-score of 0.9767. This shows that although Word2Vec efficiently detects semantic links, its performance is somewhat lower than BoW and TF-IDF, suggesting that the choice of vectorizing algorithm greatly influences the capacity of the model to distinguish between fake and true news.

%-------------combined table
\begin{table}[htbp]
% \scriptsize
\caption{Performance Metrics of SVM with Different Vectorization Techniques and Kernels}
\begin{center}
\begin{tabular}{|c|c|c|c|c|c|}
\hline
\textbf{Vectorization} & \textbf{Kernel} & \textbf{Accuracy} & \textbf{Precision} & \textbf{Recall} & \textbf{F1-Score} \\
\textbf{Technique} & & & & & \\
\hline
TF-IDF & Linear & 99.52\% & 99.53\% & 99.51\% & 0.9949 \\
\hline
Word2Vec & Linear & 96.54\% & 96.55\% & 96.53\% & 0.9644 \\
\hline
BoW & Linear & 99.81\% & 99.81\% & 99.81\% & 0.9980 \\
\hline
TF-IDF & RBF & 99.31\% & 99.32\% & 99.30\% & 0.9928 \\
\hline
Word2Vec & RBF & 97.75\% & 97.76\% & 97.74\% & 0.9767 \\
\hline
BoW & RBF & 99.62\% & 99.62\% & 99.62\% & 0.9961 \\
\hline
\end{tabular}
\label{tab:combined_results}
\end{center}
\end{table}
%-------------

Our results imply that, when combined with strong classifiers such SVM, simpler vectorizing strategies like BoW and TF-IDF can beat more sophisticated algorithms like Word2Vec in fake news identification. This is probably related to the need of word frequency in spotting false news. Though its computing cost may not be justified in many situations, the RBF kernel shown modest gains over the linear kernel in some cases, suggesting its usefulness in addressing non-linear data relationships. For identifying false news, SVM classifiers mixed with BoW or TF-IDF proved generally quite successful. Future work might investigate hybrid models using several vectorizing techniques.

%------------------------------
\subsection{Comparison with BERT Base Model}

We matched the results of our SVM model with a BERT base model trained on the same fake news dataset to further evaluate its performance. In just three epochs, the BERT base model achieved excellent results, reaching an accuracy and F1-score of 99.98\%.

\begin{table}[htbp]
\caption{Performance Metrics of BERT Base Model}
\begin{center}
\begin{tabular}{|c|c|c|}
\hline
\textbf{Epoch}  & \textbf{Accuracy} & \textbf{F1-Score} \\
\hline
3  & 0.9998 & 0.9998 \\
\hline
\end{tabular}
\label{tab:bert_results}
\end{center}
\end{table}

\begin{figure}[htbp]
\centering
\includegraphics[width=\linewidth]{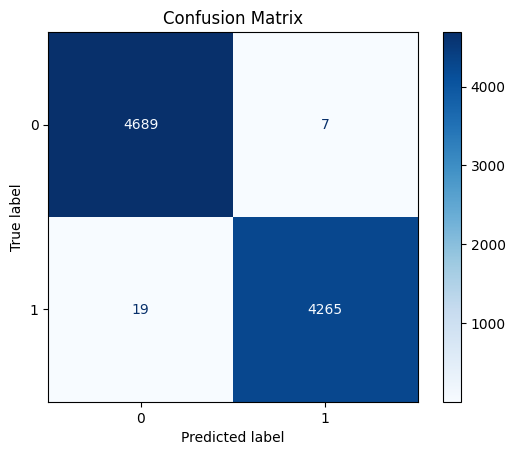}
\caption{Confusion Matrix (BERT Base)}
\label{fig:cm_bert}
\end{figure}

The confusion matrix in \textbf{Figure \ref{fig:cm_bert}} demonstrates the outstanding classification ability of the BERT base model, with only 7 false positives and 19 false negatives. The model correctly identified 4689 real news instances and 4265 fake news instances, showcasing its strong performance and ability to capture complex patterns within the dataset.

\textbf{Computational Comparison:} The BERT base model was trained using an NVIDIA Tesla T4 GPU with 16 GB VRAM, and required approximately 1 hour 45 minutes to complete 3 epochs. In contrast, the SVM model was trained on a standard CPU (Intel Core i7, 16 GB RAM) and completed training in under 3 minutes. This significant difference in training time and hardware requirements highlights the lightweight nature and computational efficiency of the SVM model, making it more suitable for scenarios with limited resources or real-time constraints.

This analysis illustrates how both SVM and BERT are effective for fake news detection. While BERT provides superior accuracy and robustness, it does so at the cost of high computational resources. On the other hand, SVM offers a faster and more efficient solution with reasonable performance. Future work could explore hybrid approaches that leverage the strengths of both models to further improve detection accuracy and robustness.

\begin{table}[htbp]

\caption{Comparison of Results with Related Work}

\begin{center}
\small
\begin{tabular}{|p{1.5cm}|p{4cm}|p{1.2cm}|p{0.8cm}|}
\hline
\textbf{Study} & \textbf{Method} & \textbf{Accuracy} & \textbf{F1-Score} \\
\hline
 \cite{zhang2018fake} & CNN + LSTM & 92.4\% & N/A \\
 \cite{perez2018automatic} & Lexical + Syntactic + Semantic & N/A & 0.78 \\
 \cite{wang2017liar} & Logistic Regression & 27.4\% & N/A \\
 \cite{ruchansky2017csi} & CSI (RNN + Credibility) & 89.2\% & N/A \\
 \cite{khattar2019mvae} & MVAE & 94.1\% & N/A \\
 \cite{bhardwaj2020fake} & CNN + BiLSTM & 96.7\% & N/A\\
\hline
\textbf{Our Work} & SVM + BoW & \textbf{99.81}\% & \textbf{0.9980} \\
\textbf{Our Work} & SVM + TF-IDF & \textbf{99.52}\% & \textbf{0.9949} \\
\textbf{Our Work} & SVM + Word2Vec & \textbf{96.54}\% & \textbf{0.9644} \\
\textbf{Our Work} & SVM (RBF) + BoW & \textbf{99.62}\% & \textbf{0.9961} \\
\textbf{Our Work} & SVM (RBF) + TF-IDF & \textbf{99.31}\% & \textbf{0.9928} \\
\textbf{Our Work} & SVM (RBF) + Word2Vec & \textbf{97.75}\% & \textbf{0.9767} \\
\textbf{Our Work} & BERT & \textbf{99.98}\% & \textbf{0.9998} \\
\hline
\end{tabular}
\label{tab:comparison}
\end{center}
\end{table}

%---------------------------conclusion with bert
\section{Conclusion}
The widespread problem of false news seriously jeopardizes the integrity of knowledge sharing and societal stability. This research has explored the efficacy of Support Vector Machines (SVM) coupled with various text vectorization methods—namely Bag of Words (BoW), Term Frequency-Inverse Document Frequency (TF-IDF), and Word2Vec—for the detection of fake news. Our comprehensive experiments demonstrate that SVM classifiers, particularly those utilizing BoW and TF-IDF, deliver outstanding performance in distinguishing fake news from genuine news. The SVM classifier with a linear kernel and BoW vectorization achieved the highest accuracy of 99.81\% and an F1-score of 0.9980, illustrating its robustness and effectiveness. The TF-IDF vectorization method also performed exceptionally well, achieving an accuracy of 99.52\% and an F1-score of 0.9949. Although the Word2Vec technique, which captures semantic relationships, showed slightly lower performance (96.54\% accuracy and 0.9644 F1-score), it remains a valuable tool in the text analysis domain.

The application of the Radial Basis Function (RBF) kernel provided marginal improvements over the linear kernel for some vectorization techniques, indicating its potential for capturing non-linear relationships in the data. However, the computational complexity associated with the RBF kernel may not be justified given the minimal performance gains observed.

Furthermore, this study compared the performance of SVM models with the BERT base model, which exhibited slightly superior performance, particularly in terms of precision and recall, achieving near-perfect scores across all metrics. Despite the BERT base model's high computational requirements, its exceptional performance underscores the potential of advanced deep learning techniques in fake news detection.

Overall, this study highlights the efficacy of straightforward vectorization techniques such as BoW and TF-IDF when paired with SVM classifiers for fake news detection. The findings suggest that robust and efficient models can be developed using these methods, contributing significantly to the automated identification of false information in news articles. 

\textbf{As a concrete direction for future work}, hybrid approaches can be explored—such as leveraging BERT to generate contextual embeddings, followed by classification using lightweight models like SVM or logistic regression. This would combine the representational power of transformer models with the computational efficiency of classical machine learning. Additionally, integrating external signals like user behavior, propagation networks, or source credibility scores may further improve detection robustness. By advancing the methodologies for detecting fake news, this study aims to support efforts in maintaining the credibility of information and curbing the spread of misinformation in the digital age.

%---------------------------

\bibliographystyle{IEEEtran}
\bibliography{references}

@article{shu2017fake,
  title={Fake news detection on social media: A data mining perspective},
  author={Shu, Kai and Sliva, Amy and Wang, Suhang and Tang, Jiliang and Liu, Huan},
  journal={ACM SIGKDD Explorations Newsletter},
  volume={19},
  number={1},
  pages={22--36},
  year={2017},
  publisher={ACM New York, NY, USA}
}

@inproceedings{zhang2018fake,
  title={Fake news detection with deep learning: A review},
  author={Zhang, Xiaoyi and Ghorbani, Ali},
  booktitle={2018 IEEE International Conference on Information Reuse and Integration (IRI)},
  pages={438--445},
  year={2018},
  organization={IEEE}
}

@inproceedings{rashkin2017truth,
  title={Truth of varying shades: Analyzing language in fake news and political fact-checking},
  author={Rashkin, Hannah and Choi, Eunsol and Jang, Jin Yea and Volkova, Svitlana and Choi, Yejin},
  booktitle={Proceedings of the 2017 conference on empirical methods in natural language processing},
  pages={2931--2937},
  year={2017}
}

@inproceedings{perez2018automatic,
  title={Automatic detection of fake news},
  author={Pérez-Rosas, Ver{\'o}nica and Kleinberg, Bennett and Lefevre, Anna and Mihalcea, Rada},
  booktitle={Proceedings of the 27th International Conference on Computational Linguistics},
  pages={3391--3401},
  year={2018}
}

@inproceedings{wang2017liar,
  title={"Liar, Liar Pants on Fire": A New Benchmark Dataset for Fake News Detection},
  author={Wang, William Yichen},
  booktitle={Proceedings of the 55th Annual Meeting of the Association for Computational Linguistics (Volume 2: Short Papers)},
  pages={422--426},
  year={2017}
}

@inproceedings{ruchansky2017csi,
  title={Csi: A hybrid deep model for fake news detection},
  author={Ruchansky, Natali and Seo, Sungyong and Liu, Yan},
  booktitle={Proceedings of the 2017 ACM on Conference on Information and Knowledge Management},
  pages={797--806},
  year={2017},
  organization={ACM}
}

@misc{FakeNewsDataset,
  author = {Clément Bisaillon},
  title = {Fake and real news dataset},
  year = {2017},
  url = {https://www.kaggle.com/clmentbisaillon/fake-and-real-news-dataset}
}

@inproceedings{khattar2019mvae,
  title={MVAE: Multimodal Variational Autoencoder for Fake News Detection},
  author={Khattar, Deepak and Goud, Jai and Gupta, Manish and Varma, Vasudeva},
  booktitle={Proceedings of the 28th ACM International Conference on Information and Knowledge Management},
  pages={141--150},
  year={2019},
  organization={ACM},
  doi={10.1145/3357384.3357817}
}

@article{bhardwaj2020fake,
  title={Fake News Detection Using Deep Learning},
  author={Bhardwaj, Amit and Bhardwaj, Amita and Pal, Harsh and Gupta, Gaurav},
  journal={International Journal of Innovative Technology and Exploring Engineering},
  volume={9},
  number={3},
  pages={2115--2119},
  year={2020},
  publisher={Blue Eyes Intelligence Engineering and Sciences Publication},
  doi={10.35940/ijitee.C8962.019320}
}

@misc{ISOT,
author = {ISOT Research Group, University of Victoria},
title = {Fake and Real News Dataset},
year = {2022},
url = {https://onlineacademiccommunity.uvic.ca/isot/2022/11/27/fake-news-detection-datasets/}
}

@article{shu2021temporal,
  title={Temporally evolving graph neural network for fake news detection},
  author={Shu, Bin Wu},
  journal={Information Processing \& Management},
  year={2021}
}

@incollection{alnabhan2024evaluating,
  title={Chapter 4 Evaluating Deep Learning for Cross-Domains Fake News Detection},
  author={Mohammad Q. Alnabhan and Paula Branco},
  booktitle={Springer Science and Business Media LLC},
  year={2024}
}

@article{verma2021welfake,
  title={WELFake: Word Embedding Over Linguistic Features for Fake News Detection},
  author={Pawan Kumar Verma and Prateek Agrawal and Ivone Amorim and Radu Prodan},
  journal={IEEE Transactions on Computational Social Systems},
  year={2021}
}

@inproceedings{kanavos2023comparative,
  title={Comparative Study of Machine Learning Algorithms and Text Vectorization Methods for Fake News Detection},
  author={Kanavos, Andreas and Karamitsos, Ioannis and Mohasseb, Alaa and Gerogiannis, Vassilis C.},
  booktitle={2023 14th International Conference on Information, Intelligence, Systems \& Applications (IISA)},
  year={2023}
}

@misc{karim2024largermodelsyieldbetter,
      title={Larger models yield better results? Streamlined severity classification of ADHD-related concerns using BERT-based knowledge distillation}, 
      author={Ahmed Akib Jawad Karim and Kazi Hafiz Md. Asad and Md. Golam Rabiul Alam},
      year={2024},
      eprint={2411.00052},
      archivePrefix={arXiv},
      primaryClass={cs.CL},
      url={https://arxiv.org/abs/2411.00052}, 
}

\end{document}